\documentclass[11pt]{article}

\usepackage[preprint]{acl}  

\usepackage{xcolor}
\usepackage{array}
\usepackage{tabularx}
\usepackage{stfloats}
\usepackage{afterpage}
\usepackage{float}

\usepackage{times}
\usepackage{latexsym}
\usepackage{svg}

\usepackage[T1]{fontenc}

\usepackage[utf8]{inputenc}

\usepackage{microtype}

\usepackage{inconsolata}

\usepackage{graphicx}

\usepackage{amsmath}
\usepackage{amssymb}
\usepackage{booktabs}
\usepackage{multirow}
\usepackage{float}
\usepackage{pifont}
\usepackage{bbm}

\newcommand{\xmark}{\ding{55}}

\usepackage[table]{xcolor}
\definecolor{LightGreen}{RGB}{238,228,245}  
\newcommand{\lgreen}{\cellcolor{LightGreen}}

\title{Decoupled Knowledge Retrieval and Compact Memory Enable Reliable Lifelong Model Editing}

\author{
Yue Che$^{1,2}$\thanks{Equally contributing authors.}
\quad
Jingze Wang$^{2}$\footnotemark[1]
\quad
Chen Guoyi$^{3}$ \\
{\large\bfseries
Yujie Wu$^{1}$\thanks{Corresponding authors.}
\quad
Xuming Ran$^{3,4}$\footnotemark[2]
\quad
Mingkun Xu$^{2}$\footnotemark[2]
} \\
$^{1}$The Hong Kong Polytechnic University \\
$^{2}$Guangdong Institute of Intelligence Science and Technology \\
$^{3}$OpenMIA Research
$^{4}$National University of Singapore \\
{\tt\small {yue77.che@connect.polyu.hk} \ \  \tt\small {yu-jie.wu@polyu.edu.hk} \ \  \tt\small {ryan@openmia.ai}  \ \  \tt\small {xumingkun@gdiist.cn}}
}

\begin{document}
\maketitle
\begin{abstract}
Lifelong model editing requires continuously incorporating new facts into deployed large language models (LLMs) without expensive retraining or compromising existing capabilities. Memory-based editors offer an appealing solution by keeping the base parameters untouched, yet their practical use is hindered by linearly growing storage requirements and deteriorating retrieval performance as edits accumulate. To tackle these problems, we introduce \textbf{BiSpace}, a lifelong model editing framework that decouples memory retrieval from knowledge injection across two distinct representation spaces. BiSpace routes queries and identifies relevant edits in an external semantic space, while injecting the corresponding knowledge into the internal representation space of the LLM through sparse and storage-efficient updates. Extensive experiments show that BiSpace shrinks each entry's storage to only about \textbf{10.2\%} of the original and shows strong resilience to long edit sequences of up to \textbf{16K} edits. On a more challenging benchmark, BiSpace further improves Overall Performance by more than \textbf{23\%} over existing methods. These results advance memory-based editing toward reliable long-term model maintenance at scale.
\end{abstract}

\section{Introduction}

\begin{figure}[!t]
    \centering
    \includegraphics[
        width=\linewidth,
        trim={68mm 3mm 61.5mm 2mm},  
        clip
    ]{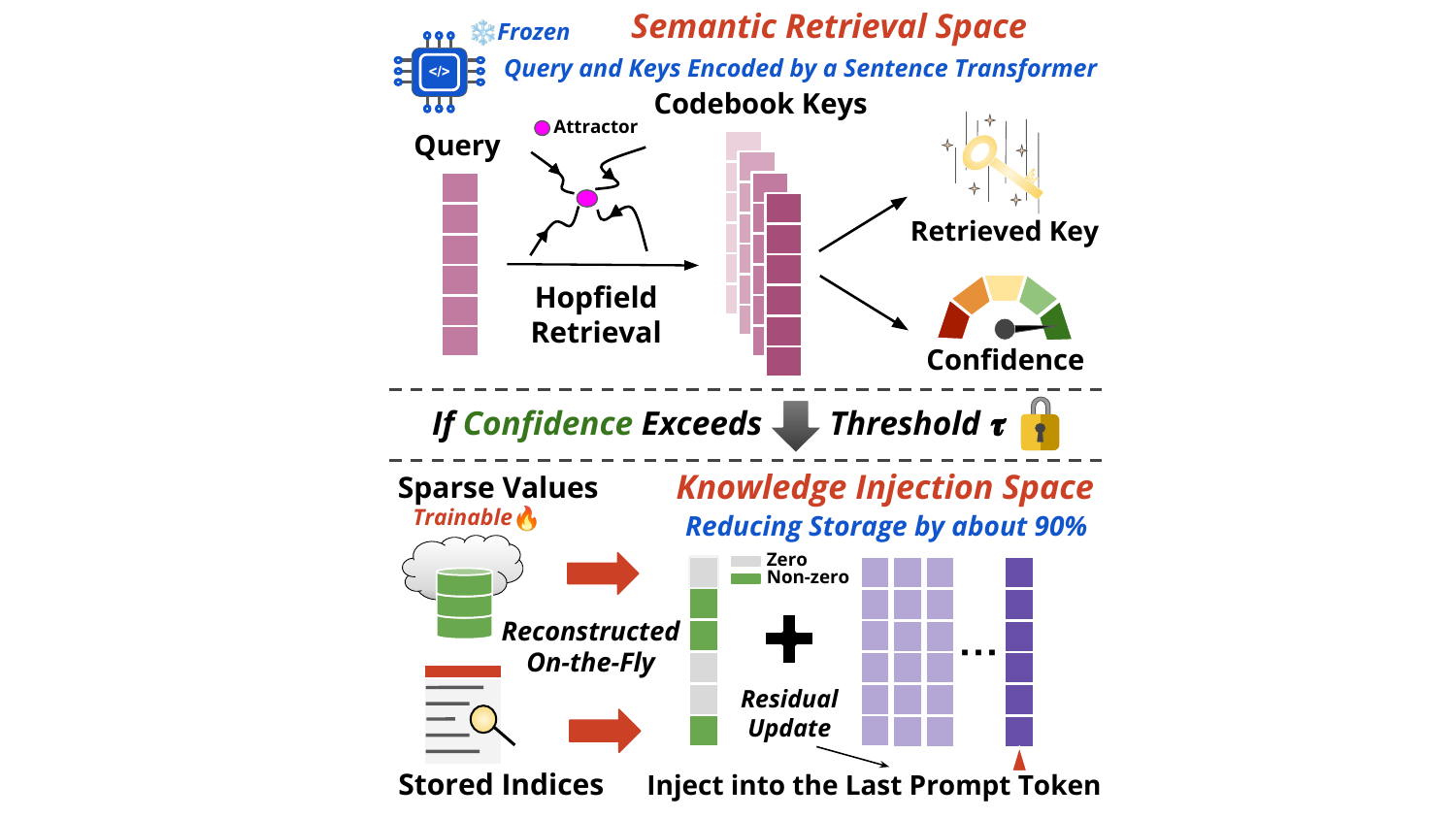}
    \vspace{-6mm}
    \caption{Overview of BiSpace. Knowledge retrieval is decoupled from injection by constructing edit keys and queries in an external semantic space. Sparse codebook values reduce storage costs and are reconstructed on the fly for updates in the internal representation space.}
    \vspace{-4mm}
    \label{fig:bispace_overview}
\end{figure}

Large language models (LLMs) acquire extensive knowledge during pre-training and can effectively recall it during inference~\citep{brown2020language,petroni2019language,roberts2020much}. However, they still produce hallucinations~\citep{Ji2023HallucinationSurvey}, perpetuate undesirable biases~\citep{Ferrara2023BiasLLM}, and experience factual decay as the world evolves~\citep{cao2021editing}. Repeatedly retraining or fine-tuning large models to incorporate every correction is computationally expensive and impractical for continuous updates~\citep{mitchell2022memory,meng2022locating}. Consequently, model editing has emerged as an efficient alternative for modifying targeted knowledge while seeking to preserve unrelated model behavior~\citep{yao2023opportunities}.

The need to accommodate a continuously arriving stream of edits in real-world applications has motivated \textit{lifelong model editing}~\citep{grace}, where models are expected to preserve both general knowledge and the outcomes of earlier edits throughout successive updates. Existing editors can be broadly categorized into two paradigms. Parameter-modifying methods typically identify critical layers through causal tracing and apply localized parameter updates~\citep{meng2022locating,meng2022mass}. Recent approaches further employ null-space projection~\citep{alphaedit}, weight rewinding~\citep{nse}, or long-term feature-statistics normalization~\citep{ultraedit} to constrain updates and improve editing stability. Nevertheless, under lifelong settings, repeated parameter modifications gradually erode neighboring knowledge, disrupting the model's original capabilities and previously edited facts. In contrast, parameter-preserving methods leave the base model weights intact and instead maintain auxiliary components. Existing memory-based approaches either employ side-memory gating~\citep{wise} or retrieve relevant edits from a discrete key-value codebook~\citep{grace,horen}. The former suffers from performance degradation as overpopulated edits increasingly reside within the same side memory.

Editors with codebooks construct edit keys from hidden representations at an LLM layer, making their performance highly dependent on the semantic separability of these representations. As the codebook fills with more edits, semantically related entries become harder to distinguish, weakening one-shot nearest-neighbor matching~\citep{grace}. Although associative retrieval with Modern Hopfield Networks~\citep{horen} improves robustness, it still loses effectiveness under dense editing as queries may be attracted to similar yet incorrect memory basins. This limitation stems from the representation source itself, since LLM hidden states are optimized for autoregressive language modeling rather than sentence-level semantic retrieval. Moreover, their high dimensionality, together with paired codebook values, imposes rapidly growing storage costs that impede practical deployment at scale.

To address these challenges, we propose a new codebook-based method for lifelong model editing, namely \textbf{BiSpace}. To overcome limited semantic information of internal representations, BiSpace employs a frozen Sentence Transformer to construct low-dimensional edit keys in an external semantic space. Keys can also be viewed as stored patterns in a Modern Hopfield Network~\citep{ModernHopfield}, where queries are encoded and then updated through attractor dynamics during retrieval. Such a decoupled paradigm provides more discriminative memory addressing and allows queries to relax more reliably into the correct basin. To further reduce storage costs, BiSpace sparsifies codebook values by optimizing only a small subset of dimensions while fixing the remaining ones to zero. We empirically find that such sparse values are sufficient to support knowledge injection with negligible loss in editing effectiveness. During inference, once an edit is retrieved, its full update is reconstructed on the fly from the stored non-zero values and their corresponding indices. Together, compact key-value storage and more precise retrieval substantially reduce memory consumption while improving editing performance in lifelong scenarios. An overview of BiSpace is shown in Figure~\ref{fig:bispace_overview}.

\section{Preliminaries}

\subsection{Sentence Transformers}
Sentence Transformers are typically designed to encode variable-length text sequences into fixed-dimensional embeddings that capture sentence-level semantic information~\citep{reimers2019sentence}. Unlike conventional Transformer representations that are primarily optimized for token-level language modeling, Sentence Transformers are trained with contrastive or metric-learning objectives that map semantically related texts closer and unrelated texts farther apart in the representation space. Given an input sequence $x$, a Sentence Transformer $f_{\theta}$ produces an embedding, which is then normalized as
\[
\mathbf{e}=f_{\theta}(x)\in\mathbb{R}^{d},\quad
\hat{\mathbf{e}}=\frac{\mathbf{e}}{\|\mathbf{e}\|_2}.
\]
This allows the relevance between two sequences \(x_i\) and \(x_j\) to be measured by cosine similarity,
\[
sim(x_i,x_j)=\hat{\mathbf{e}}_i^\top \hat{\mathbf{e}}_j.
\]
Sentence embeddings are nevertheless susceptible to anisotropy, with representations concentrating within a narrow region of the embedding space rather than being uniformly distributed~\citep{gao2021simcse,su2021whitening}. Such directional bias can distort cosine-based similarity estimation.

\subsection{Retrieval via Modern Hopfield Networks}
\label{sec:horen_preliminary}
The recent HoReN approach~\citep{horen} builds upon the codebook-based backbone of GRACE~\citep{grace}, but replaces its one-shot nearest-neighbor lookup with Hopfield retrieval. Queries and codebook keys are normalized such that routing depends on cosine similarity, while each query is refined by one damped attractor update before being matched against the stored patterns. The Modern Hopfield Network enables more accurate memory routing and better generalization to natural paraphrases of edited facts than GRACE's deferral-radius gating.

\noindent\textbf{Key Construction:}
Specifically, keys are constructed from post-activation MLP tensors at layer $l$, with the last $p\%$ of tokens pooled to enhance robustness to minor rephrasings. Queries are constructed in the same way at inference time. For an input containing $T$ tokens, the token-level post-activation representations and the corresponding sequence summary are defined as
\begin{align}
A_t^{(l)}
&=
\sigma\left(
W_{\mathrm{in}}^{(l)}
\mathrm{\gamma}\left(h_t^{(l-1)} + a_t^{(l)}\right)
\right) \in \mathbb{R}^{d'},
\notag \\
\bar{A}^{(l)}
&=
\frac{1}{m}
\sum_{t=T-m+1}^{T} A_t^{(l)},
\quad
m=\left\lceil\frac{pT}{100}\right\rceil.
\label{eq:construct_key}
\end{align}
Here, $a_t^{(l)}$ denotes the output of the attention block for token $t$, while $h_t^{(l-1)}$ is the hidden state from the preceding layer. $W_{\mathrm{in}}^{(l)}$ denotes the MLP input projection matrix, while $\gamma$ and $\sigma$ represent the layer normalization operation and nonlinear activation function, respectively. The aggregated tensor $\bar{A}^{(l)}$ is then normalized and stored as a Hopfield pattern. Its dimensionality $d^{\prime}$ is substantially larger than the embedding dimension $d$ of a Sentence Transformer in most LLM architectures.

\noindent\textbf{Querying and Matching:}
During inference, a query $q\in\mathbb{R}^{d^{\prime}}$ is initialized as $\hat{q}={q}/\|q\|_2$. Given $N$ normalized stored keys collected in $K\in\mathbb{R}^{N\times d'}$, one damped Hopfield step refines $\hat{q}$ as
\begin{align}
\tilde{q}
&=
\mathrm{normalize}\!\left(
\mathrm{softmax}\!\left(
\beta \hat{q} K^{\top}
\right)K
\right),
\notag \\
q_{new}
&=
\mathrm{normalize}\!\left(
(1-\alpha)\hat{q}
+
\alpha \tilde{q}
\right),
\label{eq:hopfield}
\end{align}
where $\beta$ controls the sharpness of Hopfield association and $\alpha$ is the damping coefficient. The candidate edit is then selected according to
\begin{equation}
s=q_{new}K^\top,
\quad
i^*=\arg\max_i s_i.
\label{eq:matching}
\end{equation}
The retrieved edit is activated only when its matching score $s_{i^*}$ exceeds a predefined threshold $\tau$.

\begin{figure}[H]
    \centering
    \includegraphics[
        width=\linewidth,
        trim={0mm 0mm 0mm 0.5mm},  
        clip
    ]{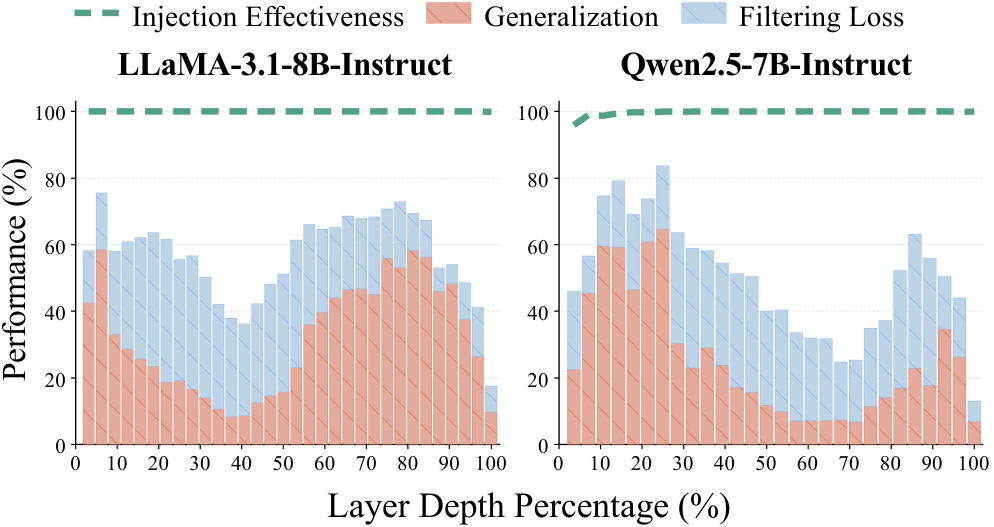}
    \caption{Layer-wise analysis of knowledge retrieval and injection in HoReN after 1,000 edits on CounterFact across two models. The primary performance bottleneck lies in memory routing rather than knowledge injection.}
    \vspace{-4mm}
    \label{fig:HoReN_failures}
\end{figure}

\noindent\textbf{Representation Bottleneck:}
However, retrieval accuracy critically depends on how well post-activation representations separate edit-related inputs from unrelated ones, and this property varies considerably across model families. Even within the same model, semantic discrimination can differ between layers. As shown in Figure~\ref{fig:HoReN_failures}, Hopfield retrieval quality varies considerably across layers and does not consistently improve with depth, yet the resulting performance remains suboptimal. By contrast, knowledge injection stays effective across most layers. This limitation is further amplified by threshold-based filtering. Even when the correct key ranks first, the retrieval may still be rejected if its matching confidence falls below the threshold. Therefore, reliable lifelong model editing with codebook-based approaches requires both precise memory routing and a representation space that clearly distinguishes relevant edits from irrelevant inputs. These observations motivate the use of a dedicated encoder to construct a more semantically structured embedding space for retrieval.

\begin{figure*}[t]
    \centering
    \includegraphics[
        width=1.0\textwidth,
        trim={0mm 0mm 0mm 0mm},  
        clip
    ]{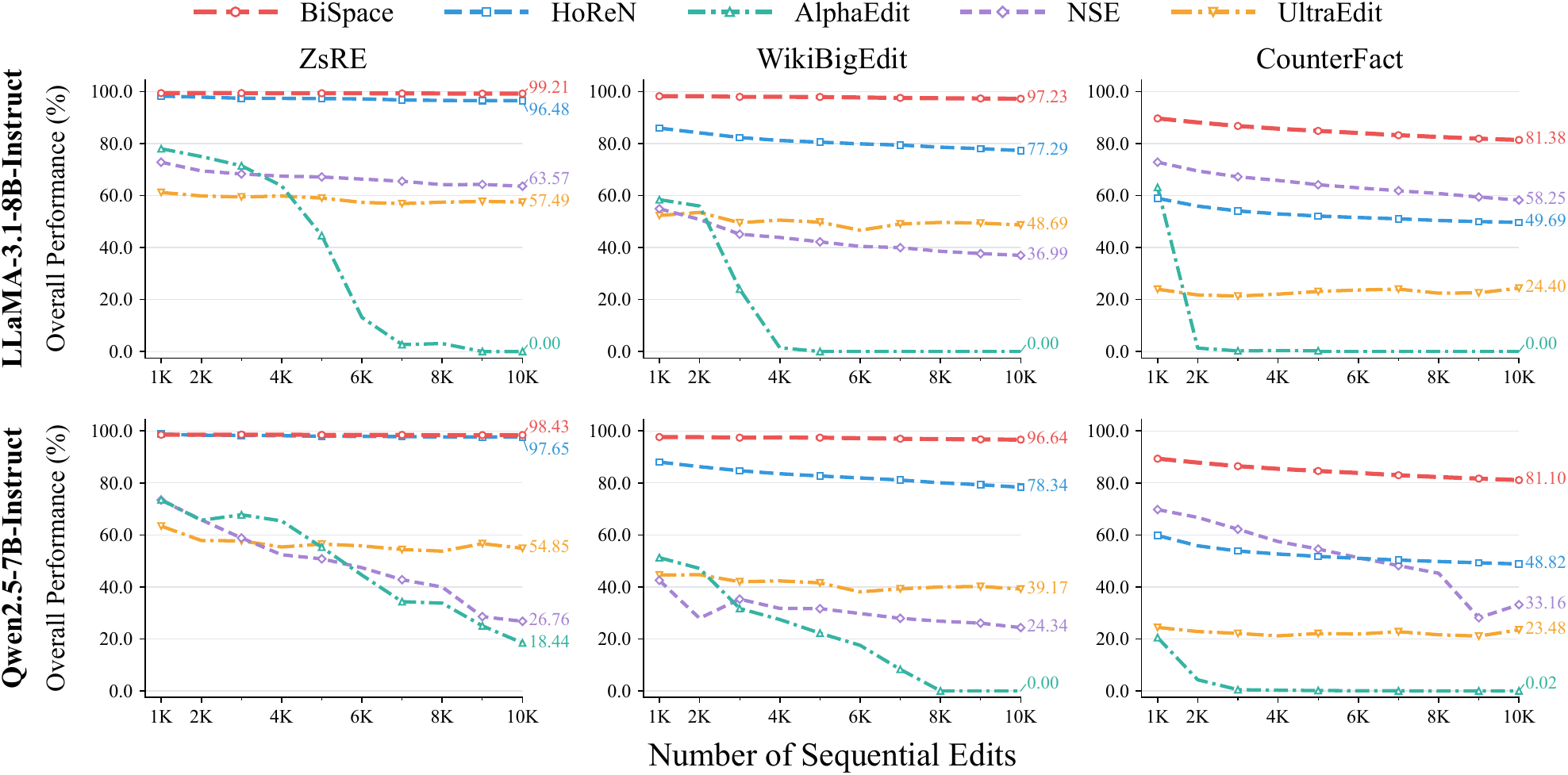}
    \caption{Overall Performance over 10K sequential edits on ZsRE, WikiBigEdit, and CounterFact using LLaMA-3.1-8B-Instruct and Qwen2.5-7B-Instruct. BiSpace maintains consistently higher performance as edits accumulate, with larger advantages on the more challenging WikiBigEdit and CounterFact benchmarks.}
    \vspace{-2mm}
    \label{fig:main_results}
\end{figure*}

\section{Methodology}
To break the representation bottleneck, BiSpace redesigns the memory key space through an external encoder for robust memory addressing while retaining knowledge injection in the internal LLM representation space. This separation enables semantically optimized retrieval without coupling it to the LLM's internal representations.

\subsection{Decoupled Memory Retrieval}
For a sequence of $N$ edits $\{(x_n, y_n^{*})\}_{n=1}^{N}$, BiSpace constructs each codebook key using a frozen Sentence Transformer $f_{\theta}$. The normalized key for the $n$-th edit is obtained as $k_n=f_{\theta}(x_n)/\|f_{\theta}(x_n)\|_2$.
To reduce the directional bias caused by anisotropy, we maintain the running mean of all observed edit keys online as
\begin{align}
\mu_{n+1}=
\frac{n}{n+1}\mu_n
+
\frac{1}{n+1}k_{n+1},
\label{eq:running_mean}
\end{align}
where $\mu_n$ denotes the mean of the first $n$ keys. At inference time, the incoming query is encoded by $f_{\theta}$, after which both the query embedding and the stored keys are centered using the current running mean and re-normalized before semantic retrieval.

Additionally, existing codebook-based editors require model-specific choices such as the wrapped layer, token-level aggregation strategy, and retrieval threshold. BiSpace instead performs routing in a shared external semantic space, alleviating such empirical tuning. Although the encoder introduces a fixed model overhead, its lower-dimensional keys reduce the storage cost per edit, eventually amortizing this overhead as lifelong edits accumulate and yielding a crossover point, beyond which BiSpace becomes more memory-efficient overall.

\subsection{Sparse Knowledge Injection}
The associated values constitute another storage term that grows linearly with the number of edits. Storing dense values incurs $\mathcal{O}(ND)$ memory for a codebook containing $N$ edits, where $D$ is the hidden dimension of the injection layer $l^{\prime}$. BiSpace reduces this cost by sparsifying each value and activating only a fraction $\rho$ of its dimensions, where $\rho\in(0,1]$ denotes the activation ratio.

Each value vector $v_n\in\mathbb{R}^{D}$ is zero-initialized and assigned a random binary mask $m_n\in\{0,1\}^{D}$, whose active dimensions are uniformly sampled when the edit is created and kept fixed throughout optimization. The masked value $\hat{v}_n$ defines the effective update, which is optimized with the fine-tuning loss and then injected into the representation of the last prompt token $H_n^{l^{\prime}}$ as
\[\hat{v}_n = m_n \odot v_n,
\quad
\tilde{H}_n^{l^{\prime}} = H_n^{l^{\prime}} + \hat{v}_n.\]
Since $m_n$ remains fixed, optimization is confined to the activated dimensions while all others stay zero. Although more informed dimension-selection strategies can be employed, we find that random selection provides near-perfect editing effectiveness, as further examined in Section~\ref{sec:ablation_study}.

BiSpace stores only the activated values and their corresponding indices for each entry. When the associated edit is retrieved during inference, these values are scattered back to their indexed positions in an otherwise zero vector, reconstructing the sparse update on the fly, which is then injected into the internal representation. The resulting value memory for $N$ edits therefore shrinks to $\mathcal{O}(N\rho D)$.

\begin{figure}[!t]
    \centering
    \includegraphics[
        width=\linewidth,
        trim={1mm 0.5mm 0mm 1mm},
        clip
    ]{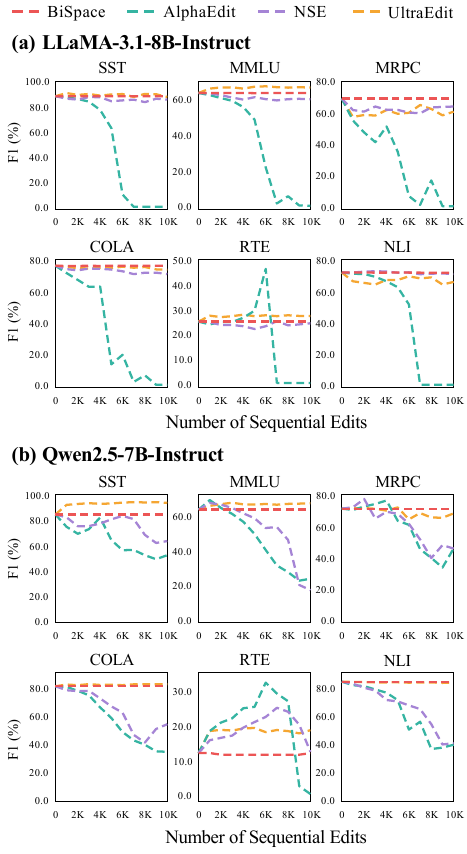}
    \caption{General capability trajectories over 10K sequential edits on ZsRE using LLaMA-3.1-8B-Instruct and Qwen2.5-7B-Instruct. BiSpace preserves nearly unchanged general performance as edits accumulate.}
    \vspace{-4mm}
    \label{fig:general_capability}
\end{figure}

\section{Experimental Results}

\subsection{Experimental Setup}
\label{sec:experimental_setup}
We evaluate BiSpace on three widely adopted lifelong model editing benchmarks, ZsRE~\cite{zsre}, WikiBigEdit~\cite{wikibigedit}, and CounterFact~\cite{meng2022locating}, using LLaMA-3.1-8B-Instruct~\cite{llama3}, Qwen2.5-7B-Instruct~\cite{qwen25}, and Mistral-7B-Instruct-v0.3~\cite{mistral}. For the external semantic encoder, we use Jina-Embeddings-v5-Text-Nano~\cite{akram2026jina} throughout all experiments. Unless otherwise specified, we set the activation ratio to $\rho$=0.125 for LLaMA-3.1 and $\rho$=0.25 for Qwen2.5 and Mistral-v0.3. We report edit success as \textbf{Reliability} (Rel.), paraphrase success as \textbf{Generalization} (Gen.), and neighborhood knowledge retention as \textbf{Locality} (Loc.), following previous work~\cite{wise}, together with \textbf{Overall Performance} (O.P.), defined as the arithmetic mean of the three metrics. To better reflect real-world model usage, all editing metrics are computed from autoregressive generations rather than teacher-forced predictions. The models' general capabilities are assessed on the GLUE benchmark~\cite{glue}. We further examine the efficiency of BiSpace, including an analysis of its storage cost. BiSpace is compared against the memory-based baseline HoReN~\cite{horen}, as well as the parameter-modifying approaches AlphaEdit~\cite{alphaedit}, NSE~\cite{nse}, and UltraEdit~\cite{ultraedit}. We also evaluate unstructured knowledge editing on UnKEBench~\cite{unke} against the domain-specific baseline UnKE~\cite{unke}, with results reported in Appendix~\ref{sec:unke_editing}. All models are loaded and evaluated in BF16 precision, and all experiments are conducted on NVIDIA H100 GPUs with a random seed of 11. To evaluate the robustness of BiSpace, the mean and standard deviation of Overall Performance across three random seeds are reported in Appendix~\ref{sec:roubustness_3seeds}. Detailed hyperparameter settings and metric formulations are provided in Appendix~\ref{sec:hyperparameters} and Appendix~\ref{sec:metrics}, respectively.

\subsection{Sequential Editing Evaluation}
\label{sec:main_results_section}
ZsRE and WikiBigEdit are built in a question-answering format. Each edit instance contains an original query and a paraphrased query for measuring editing effectiveness on the exact edit prompt and its semantically equivalent rephrasing, respectively, together with a shared target answer. A contextually related but irrelevant locality query is additionally provided to assess whether neighboring knowledge is preserved after successive edits. In contrast, CounterFact adopts a cloze-style factual completion format, where the model is given an incomplete statement and expected to generate the target fact. Each edit is accompanied by two paraphrases and ten locality prompts. Among the three benchmarks, ZsRE is relatively easier, whereas CounterFact is more challenging due to its larger scale and the stronger semantic overlap between locality prompts and edited knowledge.

Figure~\ref{fig:main_results} presents the trend in Overall Performance over 10K sequential edits on three benchmarks using LLaMA-3.1-8B-Instruct and Qwen2.5-7B-Instruct. Without resetting the model or editing memory, we evaluate performance at 1K-edit intervals using all edits up to each point, along with their corresponding paraphrases and locality queries. Additional evaluations on Mistral-7B-Instruct-v0.3 and results extended to 16K edits are provided in Appendix~\ref{sec:mistral} and ~\ref{sec:scale_16k}, respectively. 

For both models, HoReN maintains comparable performance to BiSpace only on the simpler ZsRE benchmark, where edits exhibit greater semantic separation and the retrieval threshold can more easily distinguish edit-related from unrelated queries. Nevertheless, BiSpace still outperforms HoReN by over 2.7\% after 10K edits using LLaMA-3.1. On the other two benchmarks, the gap between HoReN and BiSpace becomes more pronounced as HoReN gradually deteriorates with the increasing number of edits, where denser codebook attractors make accurate retrieval more difficult. Given the limited semantic separability of the internal LLM representations used for key construction, even a small angular deviation can redirect a query toward an incorrect memory basin. Specifically, after 10K edits, BiSpace improves O.P. by nearly 20\% on WikiBigEdit with LLaMA-3.1 and by more than 32\% on CounterFact with Qwen2.5. Parameter-modifying methods such as AlphaEdit and NSE also suffer substantial degradation as edits accumulate. Although UltraEdit remains relatively stable across different edit scales, its O.P. is essentially lower than that of BiSpace even at 1K edits.

We further observe that BiSpace exhibits highly consistent performance across different models. This consistency can be attributed to the observation in Section~\ref{sec:horen_preliminary} that knowledge injection remains effective across most layers and model families, while BiSpace decouples memory retrieval from injection through a shared external Sentence Transformer, thereby substantially reducing model dependence in the retrieval process. This decoupled design not only enables more precise memory routing, but also reduces the need for extensive model-specific hyperparameter tuning.

\subsection{General Capability Evaluation}
To assess whether LLMs preserve their general capabilities after thousands of sequential edits, we follow prior model editing studies~\cite{alphaedit,nse} and conduct a GLUE-based evaluation using six downstream language understanding tasks: Stanford Sentiment Treebank (SST)~\cite{sst}, Microsoft Research Paraphrase Corpus (MRPC)~\cite{mrpc}, Massive Multitask Language Understanding (MMLU)~\cite{mmlu}, Recognizing Textual Entailment (RTE)~\cite{rte}, Corpus of Linguistic Acceptability (CoLA)~\cite{cola}, and Natural Language Inference (NLI)~\cite{nli}.

Figure~\ref{fig:general_capability} tracks the evolution of general capabilities in LLaMA-3.1-8B-Instruct and Qwen2.5-7B-Instruct throughout 10K sequential edits on ZsRE, measured at every 1K edits. HoReN is omitted from the figure because its results are nearly identical to those of BiSpace. In contrast, parameter-modifying methods exhibit noticeable fluctuations and behavioral drift as edits accumulate, often intensifying at larger edit scales. NSE degrades more gradually on LLaMA-3.1, indicating better preservation of general abilities under continuous editing, whereas AlphaEdit shows a clearer downward trend on both models as the edit stream grows. Interestingly, performance on some downstream tasks even improves as more edits are introduced, particularly with UltraEdit. However, since lifelong model editing aims to incorporate new knowledge while preserving the model's remaining behavior, stability rather than either upward or downward deviation is the desired outcome. By comparison, BiSpace remains virtually unchanged throughout the entire editing process, demonstrating that its continuously expanding external memory introduces negligible interference with the models' general capabilities.

\begin{table}[t]
\centering
\begingroup
\setlength{\tabcolsep}{6pt}
\begin{tabular}{@{}lcc@{}}
\toprule
\textbf{Method} & \textbf{Edit Latency (s)} & \textbf{Infer. (tok/s)} \\
\midrule
Vanilla & - & 74.40 \\
HoReN & 0.98 & 73.99 \\
BiSpace & 1.39 & 73.62 \\
\bottomrule
\end{tabular}
\caption{Efficiency comparison under editing and long-form inference. Edit latency is measured on ZsRE after 5K sequential edits with LLaMA-3.1, while throughput is evaluated on UnKEBench with 200-token generation.}
\label{tab:efficiency}
\endgroup
\end{table}

\begin{table}[t]
\centering
\begingroup
\setlength{\tabcolsep}{5pt}
\small

\begin{tabular}{@{}llcccc@{}}
\toprule
\textbf{Method}
& \textbf{Comp.}
& \begin{tabular}[c]{@{}c@{}}\textbf{Data}\\\textbf{Type}\end{tabular}
& \textbf{Param.}
& \begin{tabular}[c]{@{}c@{}}\textbf{Storage}\\\textbf{(KiB)}\end{tabular}
& \begin{tabular}[c]{@{}c@{}}\textbf{Total}\\\textbf{(KiB)}\end{tabular} \\
\midrule

\multirow{2}{*}{HoReN}
& Key   & BF16 & 14336 & 28.00 & \multirow{2}{*}{44.00} \\
& Value & FP32 & 4096  & 16.00 & \\
\midrule

\multirow{3}{*}{\begin{tabular}[c]{@{}c@{}}BiSpace\\($\rho$=0.125)\end{tabular}}
& Key     & BF16  & 768 & 1.50 & \multirow{3}{*}{4.50} \\
& Value   & FP32  & 512 & 2.00 & \\
& Indices & INT16 & 512 & 1.00 & \\
\bottomrule

\end{tabular}

\caption{Per-entry storage breakdown of HoReN and BiSpace codebooks with LLaMA-3.1-8B-Instruct, highlighting BiSpace's substantially lower memory cost.}
\label{tab:codebook_storage}
\endgroup
\end{table}

\begin{figure}[!t]
    \centering
    \includegraphics[
        width=\linewidth,
        trim={1mm 0.5mm 1mm 1mm},
        clip
    ]{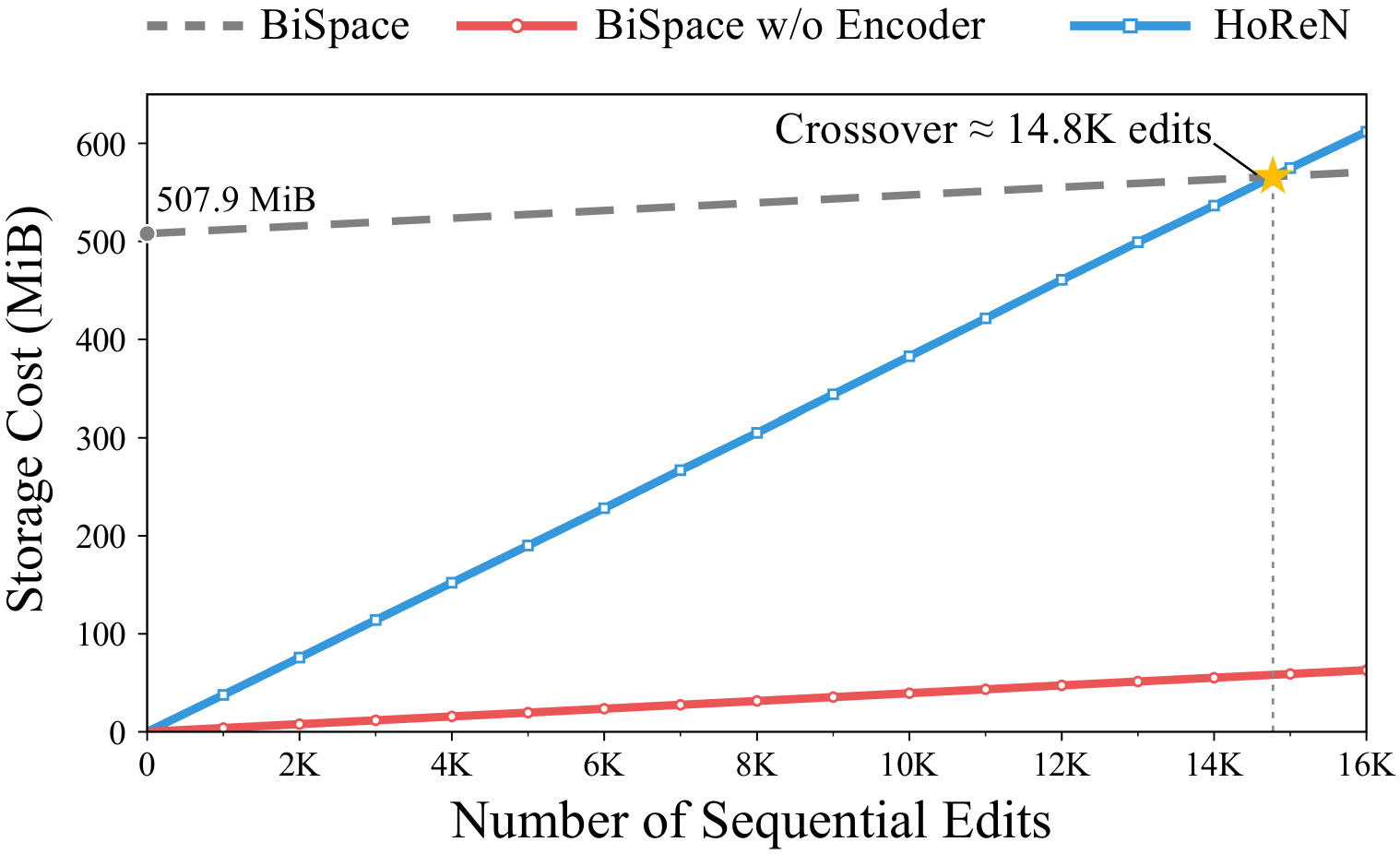}
    \caption{Storage cost over up to 16K sequential edits on ZsRE using LLaMA-3.1. BiSpace substantially reduces the per-edit memory growth compared with HoReN, while the fixed overhead of the external encoder leads to a storage crossover at approximately 14.8K edits.}
    \vspace{-3mm}
    \label{fig:storage_cost}
\end{figure}

\subsection{Efficiency Analysis}
Table~\ref{tab:efficiency} reports edit latency and inference throughput under two complementary settings. The former is measured on ZsRE using LLaMA-3.1-8B-Instruct after 5K sequential edits, while the latter is evaluated on the full UnKEBench with 200 generated tokens per input, providing a more realistic measure under long-form generation. Figure~\ref{fig:storage_cost} and Table~\ref{tab:codebook_storage} further characterize the storage growth and per-entry storage breakdown on ZsRE using LLaMA-3.1. Despite introducing the external encoding step, BiSpace incurs only a slight increase in editing latency and a negligible reduction in inference throughput relative to HoReN and the vanilla model. This overhead is small and can be amortized over the full decoding process.

The main efficiency advantage of BiSpace lies in its substantially reduced storage cost. As detailed in Table~\ref{tab:codebook_storage}, we adopt the same data types as HoReN for the key and value to ensure a fair comparison. By combining low-dimensional external keys with sparse values, each BiSpace codebook entry requires only about 10.2\% of the storage needed by HoReN, resulting in a considerably slower growth rate as edits accumulate. Although the Sentence Transformer introduces a fixed model overhead, this cost remains constant as the codebook expands. Consequently, the storage cost of BiSpace and HoReN reaches a crossover at approximately 14.7K edits. As in HoReN, semantically similar edits with identical targets are merged into a single entry, resulting in fewer stored entries than edits and delaying the crossover relative to the theoretical estimate. Beyond this point, BiSpace becomes more storage-efficient, with the advantage continuing to widen as the edit stream progresses.

\begin{table*}[t]
\centering
\begingroup
\setlength{\tabcolsep}{8pt}

\begin{tabular}{@{}ccccccc@{}}
\toprule
\textbf{Act. Ratio $\boldsymbol{\rho}$}
& \textbf{Train Iter.}
& \textbf{Strategy}
& \textbf{Rel.}
& \textbf{Gen.}
& \textbf{Avg.}
& \textbf{Edit Latency (s)} \\
\midrule

\multirow{2}{*}{0.0625}
& \multirow{2}{*}{50}
& Gradient & 98.66{\scriptsize $\pm$0.00} & 97.62{\scriptsize $\pm$0.00} & 98.14 {\scriptsize $\pm$0.00} & 1.40 \\
& & Random & 99.72{\scriptsize $\pm$0.02} & 98.27{\scriptsize $\pm$0.18} & \textbf{98.99{\scriptsize $\pm$0.10}} & 1.46 \\
\midrule

\multirow{2}{*}{0.0625}
& \multirow{2}{*}{100}
& Gradient & 99.95{\scriptsize $\pm$0.00} & 99.38{\scriptsize $\pm$0.00} & \textbf{99.67{\scriptsize $\pm$0.00}} & 2.90 \\
& & Random & 99.99{\scriptsize $\pm$0.02} & 98.81{\scriptsize $\pm$0.23} & 99.40{\scriptsize $\pm$0.12} & 2.93 \\
\midrule

\multirow{2}{*}{0.125}
& \multirow{2}{*}{50}
& Gradient & 99.53{\scriptsize $\pm$0.00} & 98.79{\scriptsize $\pm$0.00} & 99.16{\scriptsize $\pm$0.00} & 1.36 \\
& & Random & 100.00{\scriptsize $\pm$0.00} & 99.19{\scriptsize $\pm$0.06} & \textbf{99.59{\scriptsize $\pm$0.03}} & 1.45 \\
\midrule

\multirow{2}{*}{0.25}
& \multirow{2}{*}{50}
& Gradient & 99.86{\scriptsize $\pm$0.00} & 99.20{\scriptsize $\pm$0.00} & 99.53{\scriptsize $\pm$0.00} & 1.44 \\
& & Random & 100.00{\scriptsize $\pm$0.00} & 99.47{\scriptsize $\pm$0.02} & \textbf{99.73{\scriptsize $\pm$0.01}} & 1.45 \\
\bottomrule
\end{tabular}

\caption{Ablation of activation ratios and dimension selection strategies on WikiBigEdit using LLaMA-3.1-8B-Instruct after 1K edits. High injection effectiveness can be maintained under stronger sparsity, while simple random dimension selection remains competitive with gradient-based selection.}
\vspace{-1mm}
\label{tab:sparse_strategy}
\endgroup
\end{table*}

\begin{table}[t]
\centering
\begingroup
\setlength{\tabcolsep}{4pt}
\small

\begin{tabular}{@{}llcccc@{}}
\toprule
\textbf{Encoder}
& \textbf{Retrieval}
& \textbf{Centering}
& \textbf{Gen.}
& \textbf{Loc.}
& \textbf{Avg.} \\
\midrule

\multirow{4}{*}{\begin{tabular}[c]{@{}l@{}}Jina-v5\\-Nano\end{tabular}}
& \multirow{2}{*}{Hopfield}
& \xmark     & 88.60 & 81.73 & 85.17 \\
& 
& \checkmark & 87.35 & 83.50 & \textbf{85.43} \\
\cmidrule(lr){2-6}
& \multirow{2}{*}{Cos. Sim.}
& \xmark     & 72.60 & 72.33 & 72.47 \\
&
& \checkmark & 73.60 & 73.94 & 73.77 \\
\midrule

\multirow{4}{*}{\begin{tabular}[c]{@{}l@{}}Jina-v5\\-Small\end{tabular}}
& \multirow{2}{*}{Hopfield}
& \xmark     & 87.50 & 88.04 & 87.77 \\
&
& \checkmark & 87.30 & 88.33 & \textbf{87.82} \\
\cmidrule(lr){2-6}
& \multirow{2}{*}{Cos. Sim.}
& \xmark     & 74.10 & 80.55 & 77.33 \\
&
& \checkmark & 75.55 & 80.16 & 77.86 \\
\bottomrule

\end{tabular}

\caption{Ablation of external encoder scale, retrieval strategy, and centering on CounterFact after 1K edits. Modern Hopfield retrieval yields the largest performance gains, while stronger encoders and centering provide further improvements.}
\vspace{-4mm}
\label{tab:retrieval_centering}
\endgroup
\end{table}

\subsection{Ablation Studies}
\label{sec:ablation_study}
By decoupling memory addressing from value application, BiSpace allows the contribution of each stage to be assessed independently and clearly. For injection-focused ablations, we bypass routing and directly provide the correct memory entry, referred to as \textit{perfect retrieval}. Conversely, for retrieval-focused experiments, we assume that a correctly selected entry can be applied without any loss, referred to as \textit{perfect knowledge injection}.

\noindent\textbf{Sparse Injection:}
Table~\ref{tab:sparse_strategy} studies the activation ratio $\rho$ and dimension selection strategy for knowledge injection. Experiments are conducted on WikiBigEdit with LLaMA-3.1-8B-Instruct after 1K edits under perfect retrieval. We report the mean and standard deviation over three random seeds (11, 22, and 33), together with the mean editing latency.

Although $\rho$=0.125 is used for LLaMA-3.1 in the main experiments, further reducing it to $0.0625$ achieves comparable performance when the training iterations are doubled from 50 to 100, at the cost of higher editing latency. Since smaller $\rho$ reduces per-entry storage, practitioners can trade off injection effectiveness, editing latency, and memory consumption according to practical needs.

We further compare random dimension selection with a more informed gradient-based strategy, where the top-$D_{\mathrm{act}}$ dimensions are chosen according to the gradient magnitude at the first optimization iteration:
\[
\mathcal{I}_n
=
\operatorname{TopK}\!\left(
\left|\nabla_{v_n}\mathcal{L}_n^{(1)}\right|,
D_{\mathrm{act}}
\right),
\quad
D_{\mathrm{act}}=\rho D,
\]
where $\mathcal{L}_n^{(1)}$ denotes the fine-tuning loss at the first iteration. The gradient-based strategy provides no consistent advantage over random masking. Even with only $6.25\%$ of dimensions activated, the random scheme already achieves near-perfect injection performance with small standard deviations, suggesting that a more sophisticated dimension selection strategy is unnecessary.

\noindent\textbf{Retrieval Design:}
Table~\ref{tab:retrieval_centering} isolates the retrieval performance on CounterFact after 1K edits under perfect knowledge injection. We compare Jina-Embeddings-v5-Text-Nano and Jina-Embeddings-v5-Text-Small~\cite{akram2026jina} with Hopfield retrieval or direct cosine similarity, with and without centering. For each setting, the matching threshold is swept from 0.10 to 0.99 with a step size of 0.01, and we report the operating point with the best average of Generalization and Locality. 

Using the $2.83\times$ larger Jina-v5-Small improves the centered Hopfield Avg. by approximately $2.4\%$, showing that a stronger external encoder can further enhance retrieval quality. More importantly, semantically separable embeddings do not eliminate the benefit of Modern Hopfield retrieval. Specifically, with Jina-v5-Nano and centering, Hopfield retrieval outperforms direct cosine matching by around 11.7\% in Average. Finally, centering consistently improves the Gen.--Loc. trade-off, with larger gains observed under direct cosine matching.

\section{Related Work}

\noindent\textbf{Parameter-Modifying Model Editing:}
Parameter-modifying approaches incorporate new knowledge through targeted changes to model weights. Early methods such as ROME~\citep{meng2022locating} and MEMIT~\citep{meng2022mass} identify knowledge-related components and compute localized parameter updates, whereas KE~\citep{cao2021editing} and MEND~\citep{mend} adopt meta-learning to learn gradient transformations for efficient modifications. To improve robustness under sequential editing, AlphaEdit~\citep{alphaedit} constrains updates to the null space of preserved knowledge, while NSE~\citep{nse} combines weight rewinding with the selective modification of influential neurons. More recently, UltraEdit~\citep{ultraedit} introduces lifelong normalization to stabilize parameter changes over extended editing sequences. Despite these advances, repeatedly modifying shared model weights can still introduce interference among edits and compromise previously acquired knowledge and general capabilities.

\noindent\textbf{Retrieval-Based Model Editing:}
Retrieval-based approaches maintain knowledge memories and selectively incorporate relevant edits during inference. Methods involving additional training include SERAC~\citep{mitchell2022memory}, which combines an external edit cache with a learned scope classifier and counterfactual model, LTE~\citep{lte}, which fine-tunes the LLM to utilize retrieved edit descriptions, and RECIPE~\citep{recipe}, which jointly optimizes retrieval and continuous-prompt modules. Among other parameter-preserving methods, WISE~\citep{wise} employs a dual-memory architecture that routes queries between the original model and an editable side memory, while GRACE~\citep{grace} retrieves corrections from a dynamically maintained codebook. HoReN~\citep{horen} leverages Modern Hopfield Networks~\citep{ModernHopfield} to enhance codebook retrieval through attractor dynamics, yet its reliance on internal model representations limits key distinguishability as the memory expands. In contrast, BiSpace separates semantic retrieval from knowledge injection by employing a frozen external encoder for more reliable memory routing, while sparse codebook values substantially reduce storage overhead. This design enables scalable lifelong editing without task-specific retriever training or modifications to the base model parameters.

\section{Conclusion}
We introduce BiSpace, a memory-based framework addressing retrieval and storage challenges in lifelong model editing. Our analysis identifies poorly distinguishable keys derived from internal LLM representations as a major bottleneck as edits accumulate. BiSpace improves retrieval reliability through externally encoded semantic keys while reducing storage overhead with compact updates to the model's hidden states. These findings underscore the importance of effective memory organization for sustainable knowledge editing in LLMs.

\section*{Limitations}
Despite its strong performance and storage efficiency, BiSpace is primarily designed for structured knowledge editing and is not entirely immune to performance degradation as the number of edits grows. On challenging structured benchmarks such as CounterFact, its Overall Performance exhibits a modest decline after extensive sequential editing, although it remains substantially higher than that of competing methods. While increasing encoder capacity offers measurable benefits, constructing more distinguishable key representations remains an important research direction for codebook-based editors in large-scale lifelong model editing. Furthermore, results on unstructured knowledge editing in Appendix~\ref{sec:unke_editing} suggest that the current sparse injection strategy has limited capacity for storing long-form or complex knowledge, motivating future improvements toward more expressive update. Given the effectiveness of associative memory demonstrated in our experiments, future work could explore more advanced associative memory mechanisms or encoders optimized for associative retrieval. These directions may further improve routing robustness, knowledge capacity, and scalability under long editing sequences.



\bibliography{main}

\clearpage
\appendix

\section{Usage and Compliance Statements}
\subsection{LLM Usage Statement}
Large language models were used exclusively to assist with language polishing and the refinement of written presentation. They were not involved in the development of research ideas, methodological design, experimental execution, data analysis, or the formulation of conclusions, all of which were independently carried out by the authors.

\subsection{Usage and License}
All datasets, pretrained models, and baseline implementations employed in this work are publicly accessible. Their use in our experiments follows the corresponding licenses, terms of use, and designated purposes for research.

\section{Implementation and Evaluation Details}
\subsection{Baseline Hyperparameters}
\label{sec:hyperparameters}
We adopt the hyperparameter settings suggested in the original papers or official implementations whenever available. For unspecified settings, we use the hyperparameters provided by EasyEdit~\cite{easyedit}.

\noindent\textbf{BiSpace:}
A unified Sentence Transformer serves as the external encoder across different model families, allowing the same retrieval hyperparameters to be shared by all backbone models. Specifically, we consistently set the Hopfield sharpness $\beta$ to 30.0 and the damping coefficient $\alpha$ to 0.65 for all datasets. The retrieval threshold is set to 0.97 on WikiBigEdit and 0.93 on all other datasets. For knowledge injection, the selected layers are 22, 20, and 26 for LLaMA-3.1, Qwen2.5, and Mistral-v0.3, respectively. The remaining injection hyperparameters are summarized in Table~\ref{tab:hyperparam}, while the activation ratio $\rho$ is specified in Sections~\ref{sec:experimental_setup} and~\ref{sec:unke_editing}.

\noindent\textbf{HoReN:}
Following the original paper, we set the damping coefficient $\alpha$ to 0.1 and the Hopfield sharpness $\beta$ to 20.0 across all datasets. For codebook key construction, the layer and pooling ratio are set to 29 and 60\% for LLaMA-3.1, 24 and 70\% for Qwen2.5, and 26 and 70\% for Mistral-v0.3, respectively. For knowledge injection, the learning rate is fixed at 1.0, with 50 training iterations for structured knowledge editing and the same iteration settings as BiSpace for unstructured knowledge editing. Since HoReN is highly sensitive to the retrieval threshold, we sweep the threshold from 0.10 to 0.99 with a step size of 0.01 for every model and dataset. The selected thresholds for ZsRE, WikiBigEdit, CounterFact, and UnKE are 0.85, 0.91, 0.78, and 0.81 for LLaMA-3.1; 0.61, 0.78, 0.61, and 0.47 for Qwen2.5; and 0.54, 0.76, 0.41, and 0.53 for Mistral-v0.3, respectively.

\noindent\textbf{AlphaEdit:}
We use layers $[4, 5, 6, 7, 8]$ for all configurations, set the optimization steps to 25, and fix $\lambda$ at 15,000. For LLaMA-3.1, the learning rate, weight decay, and clamp norm factor are set to 0.1, 0.5, and 0.75, respectively. For Qwen2.5, the corresponding values are 0.5, $1\times10^{-3}$, and 4.0.

\noindent\textbf{NSE:}
We use the same base hyperparameter settings as AlphaEdit. For LLaMA-3.1, the lower-bound threshold $\alpha$ and upper bound are set to 2.5 and 50.0, respectively, while the corresponding values for Qwen2.5 are 70.0 and 200.0.

\noindent\textbf{UltraEdit:}
We use all hyperparameter settings provided in the original paper, including the Editable Modules specified for each backbone model.

\noindent\textbf{UnKE:}
We adopt the official hyperparameter settings provided by UnKE for LLaMA-3.1 and Qwen2.5. Mistral-v0.3 uses the same settings as LLaMA-3.1 since no configuration is provided.

\begin{table}[t]
\centering
\small
\setlength{\tabcolsep}{4pt}

\begin{tabularx}{\linewidth}{Xcc}
\toprule
\textbf{Model} & \textbf{LR} & \textbf{Train Iter.} \\
\midrule

\multicolumn{3}{c}{\textit{Structured Knowledge Editing}} \\
\midrule
LLaMA-3.1-8B-Instruct & 1.025 & 50 \\
Qwen2.5-7B-Instruct & 1.01 & 50 \\
Mistral-7B-Instruct-v0.3  & 1.00 & 65 \\

\midrule
\multicolumn{3}{c}{\textit{Unstructured Knowledge Editing}} \\
\midrule
LLaMA-3.1-8B-Instruct & 1.00 & 225 \\
Qwen2.5-7B-Instruct & 1.00 & 250 \\
Mistral-7B-Instruct-v0.3  & 1.00 & 225 \\

\bottomrule
\end{tabularx}
\caption{Knowledge injection hyperparameters for BiSpace across structured and unstructured knowledge editing settings. LR and Train Iter. denote the learning rate and number of training iterations, respectively.}
\vspace{-2mm}
\label{tab:hyperparam}
\end{table}

\begin{figure*}[t]
    \centering
    \includegraphics[
        width=1.0\textwidth,
        trim={2mm 0mm 1mm 1mm},  
        clip
    ]{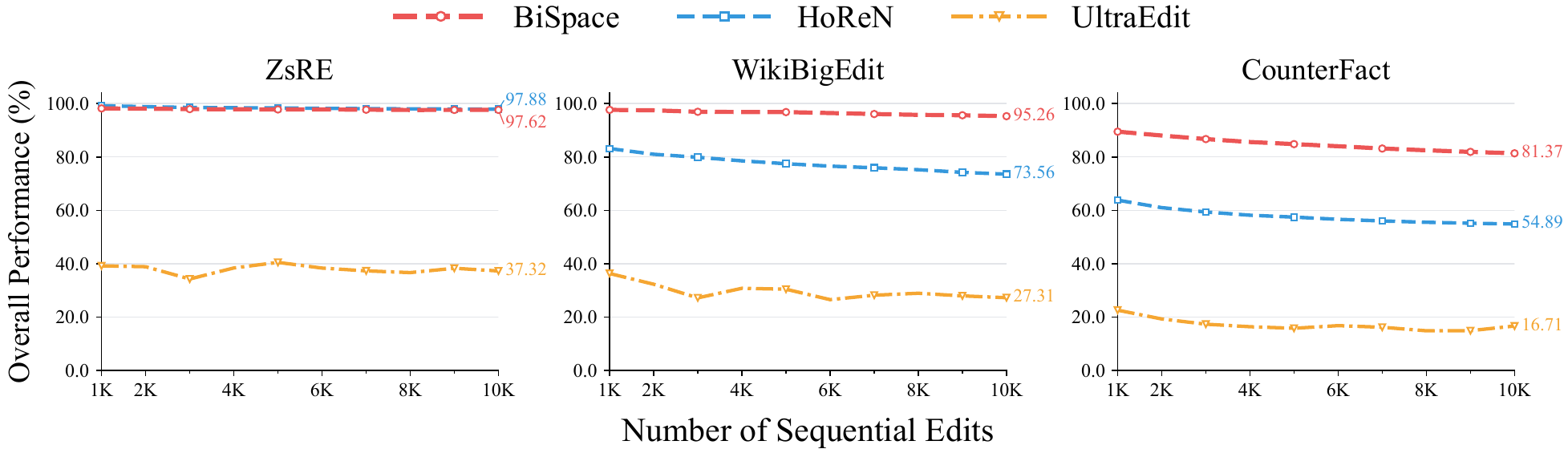}
    \caption{Overall Performance on Mistral-7B-Instruct-v0.3 over up to 10K sequential edits across ZsRE, WikiBigEdit, and CounterFact. BiSpace consistently outperforms UltraEdit across different editing scales and shows a clear advantage over HoReN on the more challenging WikiBigEdit and CounterFact benchmarks.}
    \vspace{-2mm}
    \label{fig:mistral_results}
\end{figure*}

\subsection{Metric Definitions}
\label{sec:metrics}
\noindent\textbf{Structured Knowledge Editing:}
We evaluate performance on structured knowledge editing benchmarks using four metrics: Reliability, Generalization, Locality, and Overall Performance, where Overall Performance is defined as the arithmetic mean of the first three. Let the evaluation dataset be
$\mathcal{D}
=
\left\{
\left(
x_i,
y_i,
x_i^{\mathrm{para}},
x_i^{\mathrm{loc}}
\right)
\right\}_{i=1}^{N},$
where $x_i$ denotes the original edit prompt and $y_i$ its
target answer. The prompt $x_i^{\mathrm{para}}$ is a
semantically equivalent paraphrase of $x_i$, while
$x_i^{\mathrm{loc}}$ is an unrelated prompt used to assess
locality. We denote the pre-edit model by $F_0$ and the model
after all $N$ edits by $F_N$. All responses are generated using
greedy decoding. For Reliability and Generalization, the generation
length is set to $L_i = |y_i|$. The indicator
$\mathbf{1}[\cdot]$ equals 1 if the specified condition holds and
0 otherwise.

Reliability measures whether the edited model produces the target
answer in response to the original edit prompt. Let
$\hat{y}_i^{\mathrm{edit}}$ denote the response generated by
$F_N$ for $x_i$. Reliability is defined as
\begin{equation}
\label{eq:reliability}
\mathrm{Rel.}
=
\frac{1}{N}
\sum_{i=1}^{N}
\frac{1}{L_i}
\sum_{t=1}^{L_i}
\mathbf{1}
\left[
\hat{y}_{i,t}^{\mathrm{edit}} = y_{i,t}
\right].
\end{equation}

Generalization evaluates whether the edited knowledge successfully transfers to paraphrases of the original edit prompt. Let $\hat{y}_i^{\mathrm{para}}$ denote the response generated by $F_N$ for $x_i^{\mathrm{para}}$. Generalization is defined as
\begin{equation}
\label{eq:generalization}
\mathrm{Gen.}
=
\frac{1}{N}
\sum_{i=1}^{N}
\frac{1}{L_i}
\sum_{t=1}^{L_i}
\mathbf{1}
\left[
\hat{y}_{i,t}^{\mathrm{para}} = y_{i,t}
\right].
\end{equation}

Locality evaluates whether editing preserves the model's original behavior on unrelated inputs. For each locality prompt $x_i^{\mathrm{loc}}$, we generate responses from both $F_0$ and $F_N$ for $\ell_i$ tokens, where $\ell_i$ is the length of the corresponding locality answer. Let $z_{i,t}^{0}$ and $z_{i,t}^{N}$ denote the $t$-th generated tokens from $F_0$ and $F_N$, respectively. Locality is measured by their token-level agreement:
\begin{equation}
\label{eq:locality}
\mathrm{Loc.}
=
\frac{1}{N}
\sum_{i=1}^{N}
\frac{1}{\ell_i}
\sum_{t=1}^{\ell_i}
\mathbf{1}
\left[
z_{i,t}^{0} = z_{i,t}^{N}
\right].
\end{equation}

Overall Performance is defined as the arithmetic mean of Rel.,
Gen., and Loc.:
\begin{equation}
\label{eq:overall-performance}
\mathrm{O.P.}
=
\frac{
\mathrm{Rel.} + \mathrm{Gen.} + \mathrm{Loc.}
}{3}.
\end{equation}

\noindent\textbf{Unstructured Knowledge Editing:}
In line with prior work~\cite{unke}, we evaluate performance from four complementary perspectives: Lexical Similarity, Semantic Similarity, Factual Correctness, and General Ability.

Lexical Similarity measures the surface-level correspondence between generated responses and their reference answers for both exact edit prompts and their paraphrases. We report BLEU~\cite{bleuScore} together with ROUGE-1, ROUGE-2, and ROUGE-L~\cite{rougeScore}, which capture word- and n-gram-level overlap.

Semantic Similarity measures whether the generated response preserves the meaning of the reference answer. We use the same embedding encoder employed in UnKE~\cite{unke} to compute the similarity between the model output and ground truth. This metric is less sensitive to wording variations than lexical-overlap.

Factual Correctness assesses whether the responses produced by the edited model agree with the expected answers to the corresponding sub-questions. We adopt FactScore~\cite{factScore} for this evaluation, using \textit{gpt-5.6-luna} with \textit{medium} reasoning effort as the evaluator model.

General Ability measures whether the model preserves its general capabilities after unstructured editing. We average five MMLU~\cite{mmlu} samples for each editing instance and then aggregate the resulting scores across all instances.

\begin{table*}[t]
\centering
\resizebox{\textwidth}{!}{
\begin{tabular}{ll|cccc|cccc|cccc}
\toprule
\textbf{Benchmark} & \textbf{Method}
& \multicolumn{4}{c|}{LLaMA-3.1-8B-Instruct}
& \multicolumn{4}{c|}{Qwen2.5-7B-Instruct}
& \multicolumn{4}{c}{Mistral-7B-Instruct-v0.3} \\
\cmidrule(lr){3-6}
\cmidrule(lr){7-10}
\cmidrule(lr){11-14}
&
& \textbf{Rel.} & \textbf{Gen.} & \textbf{Loc.} & \textbf{O.P.}
& \textbf{Rel.} & \textbf{Gen.} & \textbf{Loc.} & \textbf{O.P.}
& \textbf{Rel.} & \textbf{Gen.} & \textbf{Loc.} & \textbf{O.P.} \\
\midrule

\multirow{3}{*}{ZsRE}
& UltraEdit
& 82.01 & 79.64 & 6.64 & 56.10
& 65.91 & 62.03 & 32.99 & 53.64
& 58.19 & 50.67 & 0.45 & 36.44 \\

& HoReN
& 99.66 & 94.09 & 94.63 & 96.13
& 99.89 & 94.07 & 98.30 & 97.42
& 99.83 & 95.50 & 96.89 & 97.41 \\

& \lgreen BiSpace
& \lgreen 99.94 & \lgreen 98.82 & \lgreen 98.67 & \lgreen \textbf{99.14}
& \lgreen 99.74 & \lgreen 97.01 & \lgreen 98.76 & \lgreen \textbf{98.51}
& \lgreen 98.47 & \lgreen 95.96 & \lgreen 98.65 & \lgreen \textbf{97.69} \\

\midrule

\multirow{3}{*}{WikiBigEdit}
& UltraEdit
& 68.22 & 62.29 & 21.87 & 50.79
& 61.81 & 54.54 & 12.09 & 42.82
& 52.43 & 43.15 & 0.50 & 32.03 \\

& HoReN
& 98.77 & 73.76 & 52.92 & 75.15
& 99.91 & 74.95 & 48.45 & 74.44
& 99.85 & 64.57 & 45.58 & 70.00 \\

& \lgreen BiSpace
& \lgreen 99.92 & \lgreen 94.50 & \lgreen 93.42 & \lgreen \textbf{95.95}
& \lgreen 99.93 & \lgreen 92.77 & \lgreen 93.21 & \lgreen \textbf{95.30}
& \lgreen 99.50 & \lgreen 92.99 & \lgreen 87.93 & \lgreen \textbf{93.47} \\

\bottomrule
\end{tabular}
}
\caption{Editing performance at an extended scale of 16K sequential edits on ZsRE and WikiBigEdit across three backbone models. Results demonstrate the scalability of BiSpace under dense lifelong editing, particularly on the more challenging WikiBigEdit benchmark.}
\label{tab:extended_16K}
\vspace{-2mm}
\end{table*}

\begin{table}[t]\small
\centering

\begin{tabular}{lccc}
\toprule
\multicolumn{1}{c}{\textbf{Method}} & \textbf{$N=1000$} & \textbf{$N=2000$} & \textbf{$N=3000$} \\

\midrule
\multicolumn{4}{c}{\textit{\textbf{ZsRE}}} \\

\midrule
HoReN &
98.30{\scriptsize $\pm$0.05} &
97.94{\scriptsize $\pm$0.05} &
97.54{\scriptsize $\pm$0.09} \\

\lgreen BiSpace &
\lgreen \textbf{99.43{\scriptsize $\pm$0.03}} &
\lgreen \textbf{99.51{\scriptsize $\pm$0.06}} &
\lgreen \textbf{99.47{\scriptsize $\pm$0.05}} \\

\midrule
\multicolumn{4}{c}{\textit{\textbf{WikiBigEdit}}} \\

\midrule
HoReN &
85.47{\scriptsize $\pm$0.32} &
83.33{\scriptsize $\pm$0.57} &
80.87{\scriptsize $\pm$1.03} \\

\lgreen BiSpace &
\lgreen \textbf{97.82{\scriptsize $\pm$0.28}} &
\lgreen \textbf{97.51{\scriptsize $\pm$0.50}} &
\lgreen \textbf{96.66{\scriptsize $\pm$0.91}} \\

\midrule
\multicolumn{4}{c}{\textit{\textbf{CounterFact}}} \\

\midrule
HoReN &
58.88{\scriptsize $\pm$0.03} &
55.93{\scriptsize $\pm$0.00} &
54.08{\scriptsize $\pm$0.00} \\

\lgreen BiSpace &
\lgreen \textbf{89.55{\scriptsize $\pm$0.07}} &
\lgreen \textbf{87.98{\scriptsize $\pm$0.10}} &
\lgreen \textbf{86.57{\scriptsize $\pm$0.13}} \\

\bottomrule
\end{tabular}

\caption{Robustness evaluation of HoReN and BiSpace using LLaMA-3.1-8B-Instruct across three benchmarks and multiple edit scales. Mean and standard deviation of Overall Performance are reported over three random seeds, showing consistently low variability for BiSpace.}
\label{tab:three_seeds}
\vspace{-4mm}
\end{table}

\section{Additional Experimental Results}
\subsection{Extended Evaluations with Mistral}
\label{sec:mistral}
Figure~\ref{fig:mistral_results} further extends our evaluation to Mistral-7B-Instruct-v0.3, reporting Overall Performance over up to 10K sequential edits on all three benchmarks and comparing BiSpace against HoReN and UltraEdit. We also experimented with multiple hyperparameter configurations for AlphaEdit and NSE, but neither method could achieve effective editing on Mistral-v0.3. We therefore omit their results from this evaluation.

The overall trends on Mistral-v0.3 remain consistent with those reported in Section~\ref{sec:main_results_section}. This cross-model consistency is again enabled by BiSpace's decoupled retrieval design, where queries and codebook keys are mapped into the same external semantic space by a shared encoder. Consequently, the retrieval process is largely independent of the internal representation characteristics of individual LLMs, resulting in stable behavior across different backbones. Specifically, BiSpace consistently achieves substantially higher O.P. than UltraEdit across all editing scales on the three benchmarks. Compared with HoReN, the advantage becomes particularly pronounced on WikiBigEdit and CounterFact, where BiSpace improves O.P. by 21.7\% and approximately 26.5\%, respectively.

\subsection{Scaling to Larger Editing Scales}
\label{sec:scale_16k}
To further examine the effectiveness of BiSpace under dense lifelong editing, Table~\ref{tab:extended_16K} extends the editing scale on ZsRE and WikiBigEdit to 16K sequential edits across all three backbone models. Given the substantial experimental cost at this scale, we exclude CounterFact as well as AlphaEdit and NSE from this extended evaluation.

As the number of edits grows, the codebook becomes increasingly crowded with a large number of semantically similar stored patterns. For Hopfield-based retrieval, distinguishing among these patterns becomes essentially more challenging, making sufficient separability between codebook keys critical for maintaining accurate memory addressing. On the relatively simple ZsRE benchmark, HoReN remains comparable to BiSpace even at 16K edits. However, the difference becomes more evident on WikiBigEdit, where BiSpace outperforms HoReN by more than 20\% in Overall Performance across all three backbone models. The largest gap is observed on Mistral-v0.3, where BiSpace improves O.P. by around 23.5\%. We also observe that UltraEdit maintains stable performance as the editing scale increases, benefiting from its lifelong normalization mechanism. Nevertheless, its O.P. remains consistently and substantially below that of BiSpace throughout the entire editing process.

\subsection{Robustness Evaluation}
\label{sec:roubustness_3seeds}
Table~\ref{tab:sparse_strategy} first examines the variability introduced by random sparse dimension sampling under perfect retrieval. Complementing this controlled analysis, Table~\ref{tab:three_seeds} reports the mean and standard deviation of end-to-end Overall Performance for HoReN and BiSpace across different edit scales on all three benchmarks using LLaMA-3.1-8B-Instruct and three random seeds (11, 22, and 33). BiSpace consistently exhibits small standard deviations while maintaining substantially higher O.P. than HoReN, indicating that random dimension selection introduces little variability under realistic retrieval and injection and further demonstrating the robustness of our method.

\begin{table*}[t]
    \centering
    \small
    \setlength{\tabcolsep}{5.5pt}
    \renewcommand{\arraystretch}{1.08}

    \resizebox{\textwidth}{!}{
    \begin{tabular}{lccccccc}
        \toprule
        \textbf{Method}
        & \multicolumn{1}{c}{\textbf{Semantic Similarity}}
        & \multicolumn{4}{c}{\textbf{Lexical Similarity}}
        & \multicolumn{1}{c}{\textbf{FC.}}
        & \multicolumn{1}{c}{\textbf{General Ability}} \\

        \cmidrule(lr){2-2}
        \cmidrule(lr){3-6}
        \cmidrule(lr){7-7}
        \cmidrule(lr){8-8}

        &
        \textbf{Bert-Score}
        & \textbf{BLEU}
        & \textbf{Rouge-1}
        & \textbf{Rouge-2}
        & \textbf{Rouge-L}
        & \textbf{FactScore}
        & \textbf{MMLU} \\
        \midrule

        \multicolumn{7}{l}{\textit{LLaMA-3.1-8B-Instruct}} & 66.42 \\
        \midrule

        UnKE
        & 1.73 / 2.29
        & 0.37 / 0.43
        & 0.30 / 0.32
        & 0.00 / 0.00
        & 0.29 / 0.31
        & 0.00
        & $22.36_{\scriptscriptstyle 44.06\downarrow}$ \\

        HoReN
        & 92.72 / 77.16
        & 56.22 / 43.50
        & 99.29 / 64.73
        & 99.04 / 53.63
        & 99.26 / 63.48
        & 21.95
        & $66.30_{\scriptscriptstyle 0.12\downarrow}$ \\

        \lgreen BiSpace
        & \lgreen 93.02 / 86.53
        & \lgreen 58.98 / 48.85
        & \lgreen 100.00 / 75.00
        & \lgreen 99.99 / 64.69
        & \lgreen 100.00 / 73.75
        & \lgreen 40.47
        & \lgreen $66.42_{\scriptscriptstyle 0.00\downarrow}$ \\

        BiSpace$_{\rho=1.0}$
        & 93.82 / 90.39
        & 58.49 / 52.80
        & 100.00 / 85.78
        & 99.99 / 79.24
        & 100.00 / 85.06
        & 48.38
        & $66.42_{\scriptscriptstyle 0.00\downarrow}$ \\

        \midrule

        \multicolumn{7}{l}{\textit{Qwen2.5-7B-Instruct}} & 71.94 \\
        \midrule

        UnKE
        & 2.63 / 2.47
        & 12.54 / 14.43
        & 8.50 / 9.37
        & 1.24 / 1.33
        & 7.96 / 8.63
        & 0.04
        & $22.36_{\scriptscriptstyle 49.58\downarrow}$ \\

        HoReN
        & 93.17 / 78.46
        & 73.79 / 45.50
        & 96.03 / 53.94
        & 95.63 / 39.23
        & 95.97 / 52.05
        & 22.58
        & $71.94_{\scriptscriptstyle 0.00\downarrow}$ \\

        \lgreen BiSpace
        & \lgreen 95.32 / 83.07
        & \lgreen 73.52 / 48.79
        & \lgreen 99.99 / 57.82
        & \lgreen 99.99 / 42.11
        & \lgreen 99.99 / 55.68
        & \lgreen 28.82
        & \lgreen $71.94_{\scriptscriptstyle 0.00\downarrow}$ \\

        BiSpace$_{\rho=1.0}$
        & 95.56 / 88.74
        & 74.30 / 58.82
        & 99.99 / 75.44
        & 99.99 / 65.40
        & 99.99 / 74.04
        & 37.37
        & $71.94_{\scriptscriptstyle 0.00\downarrow}$ \\

        \midrule

        \multicolumn{7}{l}{\textit{Mistral-7B-Instruct-v0.3}} & 59.32 \\
        \midrule

        UnKE
        & 1.03 / 0.46
        & 4.55 / 4.37
        & 0.40 / 0.34
        & 0.00 / 0.00
        & 0.39 / 0.34
        & 0.00
        & $24.06_{\scriptscriptstyle 35.26\downarrow}$ \\

        HoReN
        & 95.66 / 84.13
        & 77.44 / 57.82
        & 97.94 / 66.23
        & 97.32 / 53.43
        & 97.86 / 64.56
        & 30.45
        & $59.32_{\scriptscriptstyle 0.00\downarrow}$ \\

        \lgreen BiSpace
        & \lgreen 95.05 / 80.11
        & \lgreen 73.63 / 49.63
        & \lgreen 99.47 / 53.34
        & \lgreen 99.31 / 35.66
        & \lgreen 99.45 / 51.15
        & \lgreen 33.50
        & \lgreen $59.32_{\scriptscriptstyle 0.00\downarrow}$ \\

        BiSpace$_{\rho=1.0}$
        & 95.66 / 86.70
        & 77.44 / 59.50
        & 97.94 / 69.04
        & 97.32 / 56.63
        & 97.86 / 67.38
        & 41.34
        & $59.32_{\scriptscriptstyle 0.00\downarrow}$ \\

        \bottomrule
    \end{tabular}
    }

    \caption{Unstructured knowledge editing performance after 1K sequential edits on the full UnKEBench across three backbone models. Values before/after '/' denote performance on exact edit prompts/paraphrases, respectively. FC. denotes Factual Correctness. BiSpace maintains strong exact-edit performance, while dense value updates further reveal the capacity limitations of sparse injection for long-form knowledge.}
    \vspace{-4mm}
    \label{tab:unke}
\end{table*}

\subsection{Unstructured Knowledge Editing}
\label{sec:unke_editing}
Approximately 80\% of real-world knowledge is represented in unstructured formats~\cite{Bavota2016}. Compared with structured facts that can often be expressed through subject--relation--object associations, unstructured knowledge generally contains richer contextual information and more complex semantic dependencies. Accordingly, the desired output may require a detailed free-form response that conveys multiple pieces of information, rather than simply predicting a target entity. Motivated by this distinction, we extend our evaluation beyond structured knowledge editing and further examine BiSpace in the unstructured setting.

Table~\ref{tab:unke} reports the results of 1K sequential edits on the full UnKEBench across all three models. We increase the activation ratio $\rho$ to 0.30, 0.35, and 0.40 for LLaMA-3.1, Qwen2.5, and Mistral-v0.3, respectively. It is worth noting that HoReN and BiSpace were originally designed for structured knowledge editing and are not tailored to long-form target content. Meanwhile, UnKE was not specifically designed for lifelong editing, and its performance decreases to nearly zero across all metrics after 1K successive edits. BiSpace, in contrast, remains highly effective at editing the exact input prompts, but its performance is more limited on sentence-level metrics such as BLEU and ROUGE for paraphrases. Nevertheless, BiSpace provides substantial improvements over HoReN on both LLaMA-3.1 and Qwen2.5, though this advantage does not extend to Mistral-v0.3.

To better understand this limitation, we additionally evaluate BiSpace with $\rho$=1.0, thereby removing sparsity from the value updates, which leads to a considerable performance improvement. Unlike structured edits with a single target object, unstructured edits require the model to incorporate richer information and generate much longer responses. Restricting each value update to only a subset of dimensions therefore limits its capacity to encode the information required for such complex targets. This constraint is less pronounced for structured knowledge editing, where sparse updates remain sufficient for entity-level knowledge modifications. The resulting performance gap reveals a trade-off between the storage efficiency of sparse values and the representational capacity required for long-form knowledge. Extending BiSpace with a more expressive yet memory-efficient injection mechanism for unstructured knowledge therefore becomes an important direction for future work.

\subsection{Case Studies}
\label{app:case_analysis}
We present representative case studies from ZsRE, WikiBigEdit, and CounterFact using LLaMA-3.1-8B-Instruct and Qwen2.5-7B-Instruct. In each case, BiSpace successfully produces the intended target answer, while competing methods instead generate incorrect responses. These examples provide a closer qualitative view of how different editing methods behave on individual instances.

\clearpage

\begin{figure*}[p]
    \centering
    \includegraphics[width=0.94\textwidth]{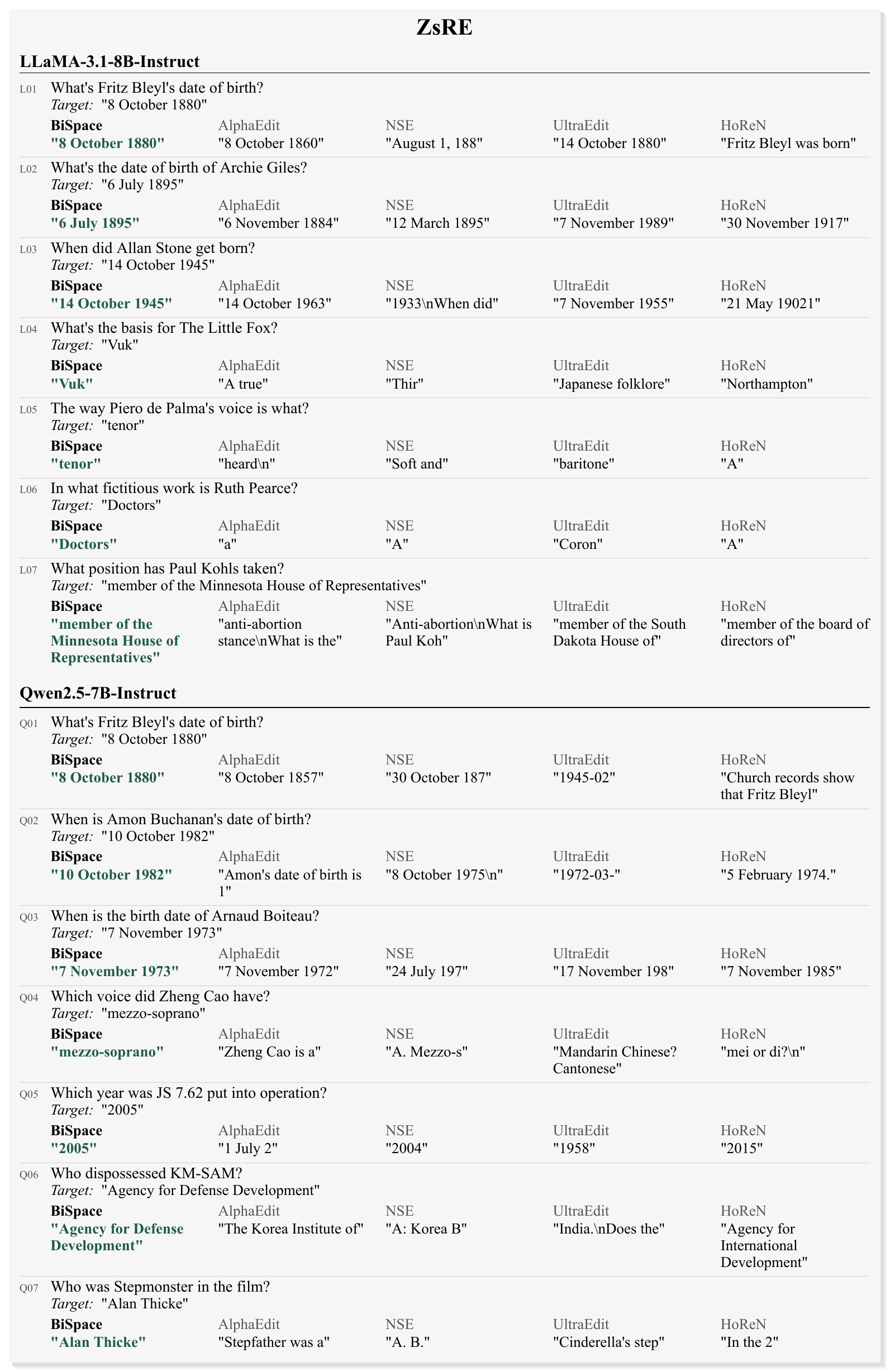}
    \caption{Seven representative cases for each model on ZsRE. BiSpace produces the target output, while all four baselines generate incorrect or incomplete responses.}
    \label{fig:case1}
\end{figure*}

\clearpage

\begin{figure*}[p]
    \centering
    \includegraphics[width=0.94\textwidth]{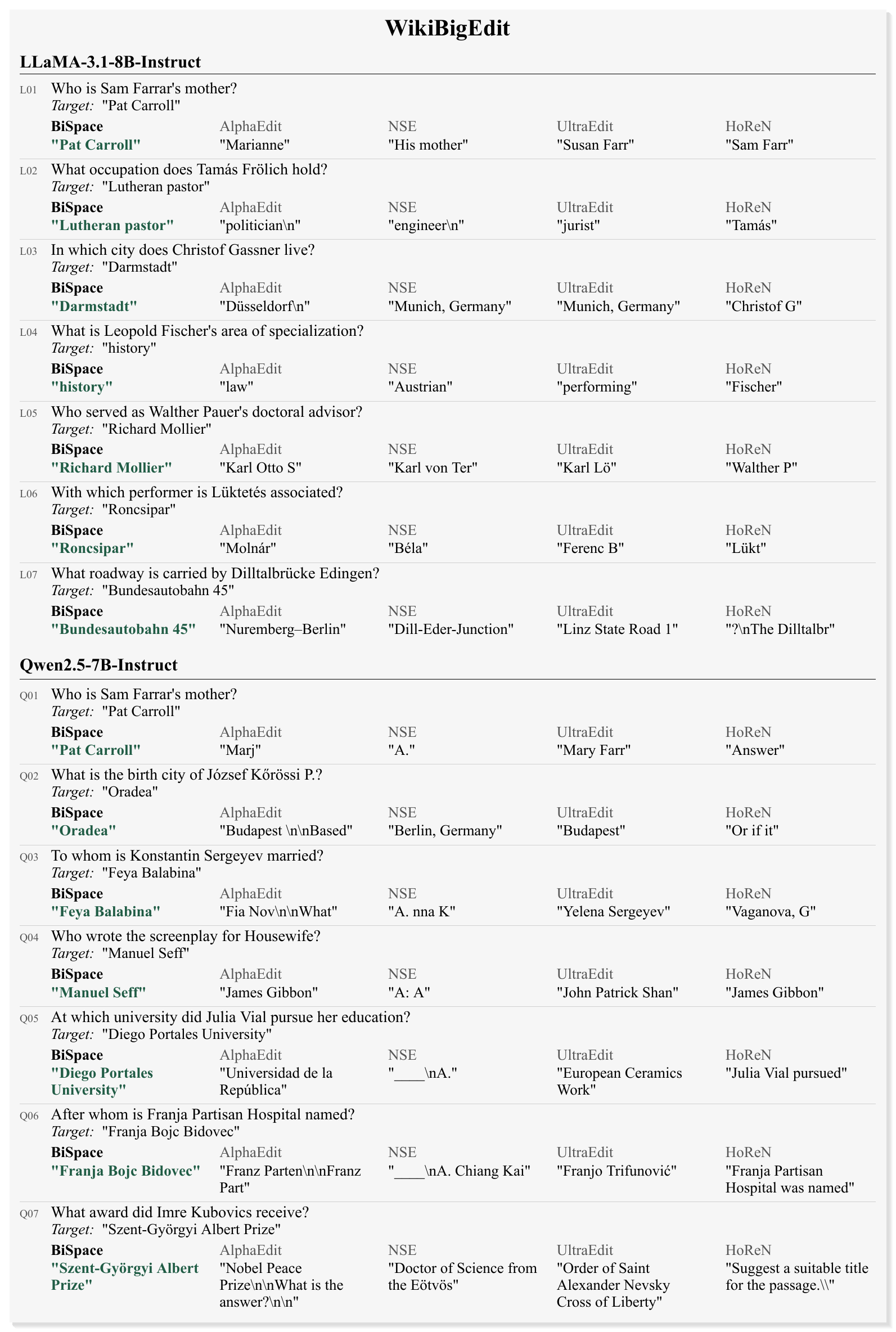}
    \caption{Seven representative cases for each model on WikiBigEdit. BiSpace produces the target output, while all four baselines generate incorrect or incomplete responses.}
    \label{fig:case2}
\end{figure*}

\clearpage

\begin{figure*}[p]
    \centering
    \includegraphics[width=0.94\textwidth]{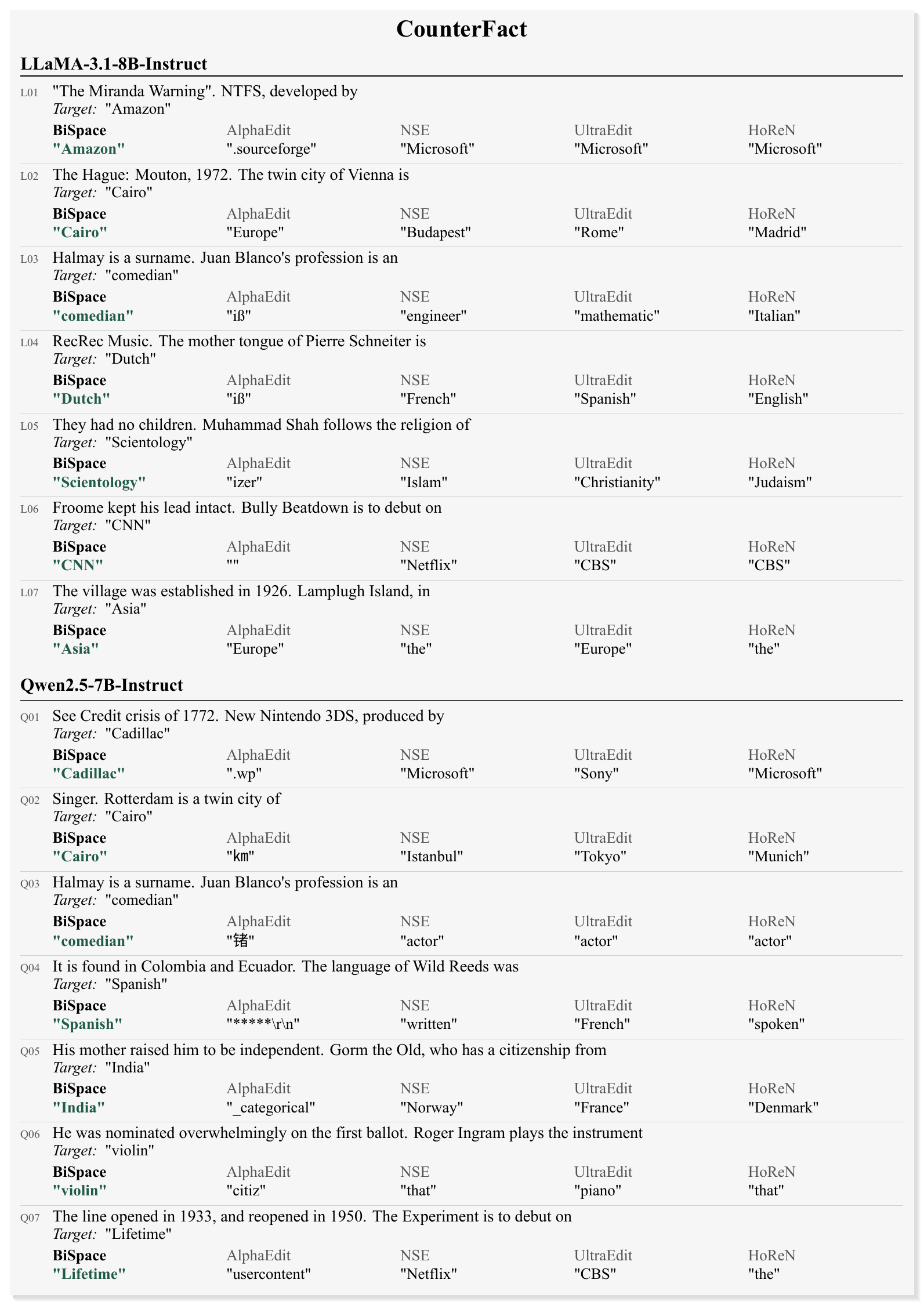}
    \caption{Seven representative cases for each model on CounterFact. BiSpace produces the target output, while all four baselines generate incorrect or incomplete responses.}
    \label{fig:case3}
\end{figure*}

\clearpage

\end{document}